\documentclass[letterpaper,10pt,conference]{ieeeconf}

\IEEEoverridecommandlockouts                              % This command is only needed if 
\usepackage{graphicx} % Required for inserting images
\usepackage{xcolor}
\usepackage{algorithm}
\usepackage{algorithmic}
\usepackage{amsmath}
\usepackage{amssymb}
\usepackage{cite}
\usepackage{booktabs}
\usepackage{stfloats}
\usepackage{svg}
\usepackage{makecell}
\usepackage{subcaption}
\usepackage[font=small]{caption}
\usepackage{hyperref}

\newcommand{\Rtheta}{R_b^i(\boldsymbol{\theta})}
\newcommand{\pos}{\boldsymbol{p}}
\newcommand{\vel}{\boldsymbol{v}}
\newcommand{\accel}{\boldsymbol{a}}
\newcommand{\att}{\boldsymbol{\theta}}
\newcommand{\angrates}{\boldsymbol{\omega}}
\newcommand{\gyrobias}{\boldsymbol{b}_{\text{gyro}}}
\newcommand{\state}{\boldsymbol{x}}
\newcommand{\inputs}{\boldsymbol{u}}
\newcommand{\zeromatrix}{\boldsymbol{0}_{3\times3}}

\title{ROScopter: A Multirotor Autopilot based on ROSflight 2.0}
\author{Jacob Moore$^1$,  Ian Reid$^1$, Phil Tokumaru$^2$, Randy Beard$^1$, Tim McLain$^1$% <-this % stops a space
\thanks{$^{1}$Brigham Young University}
\thanks{$^{2}$AeroVironment Inc.}
}
\date{\today}

\begin{document}

\maketitle

\begin{abstract}
    ROScopter is a lean multirotor autopilot built for researchers.
    ROScopter seeks to accelerate simulation and hardware testing of research code with an architecture that is both easy to understand and simple to modify.
    ROScopter is designed to interface with ROSflight 2.0 and runs entirely on an onboard flight computer, leveraging the features of ROS 2 to improve modularity.
    This work describes the architecture of ROScopter and how it can be used to test application code in both simulated and hardware environments.
    Hardware results of the default ROScopter behavior are presented, showing that ROScopter achieves similar performance to another state-of-the-art autopilot for basic waypoint-following maneuvers, but with a significantly reduced and more modular code-base.
\end{abstract}

\section{Introduction} \label{introduction}
Unmanned aerial vehicles (UAVs) have gained significant popularity in recent years due to applications such as search-and-rescue, photography, package delivery, and advanced air mobility.
Advances in autopilot technology have lowered the cost of UAV hardware, increasing availability for researchers.

Successful UAV research requires significant simulation and hardware testing of application code to ensure correct and safe functionality.
Additionally, to test this application code, UAV researchers often need low-level access to the internals of the autopilot.
While many excellent open-source autopilots exist \cite{px4,ardupilot}, integration of application code into such autopilots can be difficult and time-intensive due to large feature sets and significant portions of code residing on embedded computers.

ROSflight\cite{rosflight2025} is a lean, research-focused autopilot designed to accelerate testing of application code in both simulation and hardware.
The ROSflight firmware runs on an embedded flight controller unit (FCU), interfacing directly with sensors and actuators on the physical aircraft, while most of the autonomy stack is offloaded to an onboard companion computer.
The onboard computer communicates with the FCU over a high-speed serial connection.

ROScopter is an autonomy stack for multirotor vehicles that is designed to interface with ROSflight firmware.
ROScopter is lean, and therefore does not boast the impressive feature set of other open-source autopilots like PX4\cite{px4} or ArduPilot\cite{ardupilot}, instead offering only basic waypoint-following functionality out of the box.
While this places more responsibility on a user to develop application-specific code and necessary features, this lean feature set reduces the size of ROScopter's code-base and increases understandability.
Increasing understandability lowers the barrier to entry for UAV researchers and students, and can reduce the time required to implement and test application code in both simulation and hardware environments.
ROScopter is built around the Robot Operating System (ROS 2)\cite{ros2_2022} and leverages the features of ROS 2 to enhance modularity.

The ROScopter project is open-source and focuses on clean code and complete documentation.
Links to the GitHub codebase and the documentation can be found on the project website, \href{rosflight.org}{rosflight.org}.

In this paper we present a novel ROS 2-based autonomy stack for multirotors, intended to be used with ROSflight.
We describe the intended use case for ROScopter and show how it can be used to accelerate UAV research,
The performance and modularity of ROScopter is demonstrated with simulation and hardware results.
Finally, we compare ROScopter to a state-of-the-art open-source autopilot in a waypoint-following mission.

\section{Related Work} \label{related-work}
As UAVs have gained popularity, many excellent autopilots have been developed.
Open-source projects like PX4\cite{px4} and ArduPilot\cite{ardupilot} have been used extensively for research, commercial, and hobbyist applications.
These projects have extensive community support and offer strong plug-and-play functionality with extensive feature sets that work out-of-the-box.
While useful in many applications, these large feature sets lead to large code-bases and a black-box environment for users unfamiliar with the code. 
This can decrease understandability and increase the learning curve \cite{opensource_autopilot_survey}, both of which increase researcher effort.
Additionally, when errors do occur, the black-box environment makes debugging difficult and time-intensive.

Both PX4 and ArduPilot support many different airframes and vehicle configurations.
However, when non-standard aircraft (e.g. tilt-rotor) are desired, many researchers opt to write firmware from scratch \cite{px4_custom_controller}.
To avoid writing custom firmware from scratch for novel controllers, the authors in \cite{px4_custom_controller} develop a template to more easily integrate custom controllers into PX4.
This template approach can reduce researcher effort when integrating custom controllers, but does not reduce the complexity or learning curve of the autopilot.
ROScopter offers a reduced feature set, that decreases the size of the code-base and improves understandability of the autopilot.

Autonomy stacks for most autopilots run on embedded computers.
Autopilots such as  PX4 and ArduPilot have excellent support for onboard computer control, enabling users to develop application code in a Linux-based environment.
Despite excellent support for high-level control from an onboard computer, low-level control, like direct motor commands for a multirotor, is not well-supported \cite{PX4-external-modes-limitations}.
ROScopter moves the entire autonomy stack to the companion computer, including estimation, path planning, and control loops.
This allows users to develop application code in a Linux-based environment regardless of whether users write high or low-level control and estimation loops.

Recently, PX4 has increased support for ROS 2, further improving modularity\cite{opensource_autopilot_survey}.
ROScopter improves this modularity by wrapping every module in the autonomy stack in a ROS 2 node to allow for runtime configuration of the entire autonomy stack.
This also enables researchers to use powerful ROS 2 introspection and debugging features to assist development in all portions of the autonomy stack.

Other autopilots like \cite{betaflight} or \cite{inav} have large hobbyist communities, but are less suited to UAV research due to limited support for onboard companion computers.
Additionally, these projects do not have any official support for ROS 2.

ROScopter is designed to fulfill the needs of researchers, students, and early-stage development projects, not production-ready applications.
Thus, it does not support all of the features or functionality present in other state-of-the-art autopilots like PX4 or ArduPilot, instead focusing on understandability, modularity, and ease of integration from simulation to hardware.

\section{ROScopter Architecture}\label{system-architecture}
\begin{figure}
    \centering
    \includegraphics[width=0.9\linewidth]{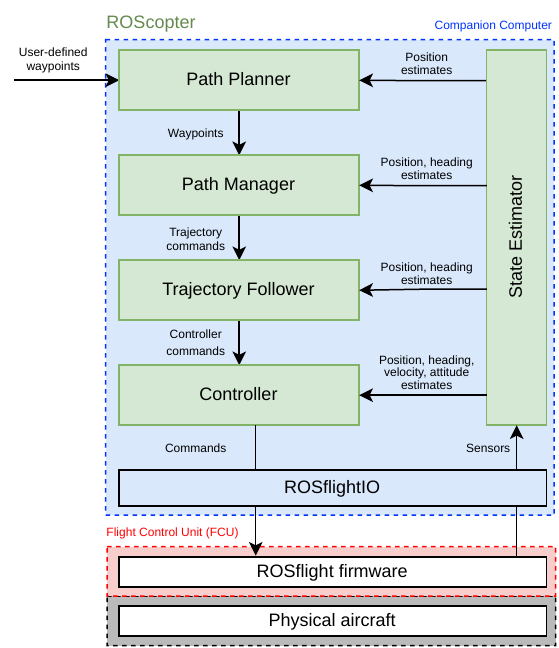}
    \caption{Architecture of ROScopter within the ROSflight framework. ROScopter communicates with the ROSflight firmware via the ROSflightIO node.}
    \label{fig:roscopter-system-architecture}
    \vspace{-15pt}
\end{figure}

The ROScopter autonomy stack has a cascading architecture shown in Figure \ref{fig:roscopter-system-architecture}.
High-level navigation modules perform path planning and trajectory-following functions, while the controller manages lower-level commands.
A full-state extended Kalman filter performs state estimation and provides the estimated state to each module.
ROScopter is designed to interface with ROSflightIO, which is a ROS 2-based I/O node that communcates over a serial connection with the ROSflight flight control unit (FCU) \cite{rosflight2020}.
The following sections first describe the structure of the ROScopter code, and then each module is described in detail.

\subsection{Code Structure}\label{system-architecture:code}
Each module in ROScopter is a ROS 2 node with a single responsibility, and takes advantage of ROS 2 interfaces (e.g. publishers, subscribers and services) to communicate with other nodes.
This separation enhances the modularity of the software and enables users to modify ROScopter with low effort, since each node can be modified independently of all other nodes.
ROS 2 also enables easy runtime configuration of the autonomy stack, as users can swap nodes or configure parameters at runtime from a launch file.
Using ROS 2 to communicate between modules also enables seamless cross-language compatibility between modules.

Each module in ROScopter follows an inheritance pattern to facilitate modifying and extending the modules.
In this architecture, a base class defines all of the ROS 2 interfaces associated with the node, in addition to virtual \emph{work} functions.
These virtual work functions are implemented by a derived class and implement the actual functionality, or work, performed by the node.
For example, the trajectory follower interface class defines its ROS 2 subscribers and publishers (e.g. commanded trajectory and low-level controller command topics, respectively).
The work function required by the trajectory follower interface class is responsible for computing the controller commands to follow the received trajectory commands.
Thus, changing how the trajectory follower computes those controller commands simply requires a user to inherit from the the interface class and implement the required work function.
Since the ROS 2 interfaces remain the same, the new code interfaces seamlessly with the rest of the ROScopter framework.

\subsection{State Estimation}
ROScopter estimates the position, velocities, attitude and gyroscope biases of the multirotor.
The state vector $\state$ is,
\begin{equation} \label{eq:state_vector}
\state = 
\begin{bmatrix}
\pos^\top & \vel^\top & \att^\top & \gyrobias^\top
\end{bmatrix}^\top.
\end{equation}

The position of the aircraft, $\pos$, is expressed in the inertial frame using NED coordinates in meters.
The velocity of the body frame, $\vel$, is expressed in the body-frame in meters per second.
The attitude of the aircraft, $\att$, relative to the inertial frame in radians using the Euler angles roll, pitch and yaw, $\phi,\,\theta,\,\psi$ respectively.
The Euler angles use the ZYX convention of application as in \cite{uavbook}.
Finally, $\gyrobias$ is the bias in the gyroscope measurements expressed in body-frame coordinates in radians per second.
The attitude is modeled using Euler angles rather than a quaternion or other Lie group-based orientation representation for ease of interpretability.
A major focus of ROScopter is understandability and extensibility and an Euler formulation helps facilitate the understanding of the estimator.
In addition, the results show that the performance is comparable to a quaternion-based estimation despite the suboptimality of an Euler-angle formulation for the flight regimes that ROScopter is designed for.
Furthermore, ROScopter's modularity allows for more advanced estimators to be easily integrated if necessary for a researcher's application.

\subsubsection{State Propagation}
The state estimator is a continuous-discrete formulation as in \cite{uavbook}.
This has the advantage of having a direct connection to the familiar dynamical equations and a clear relationship to the discrete time formulation that follows.

The dynamics of the aircraft state are modeled as a function of the state and the input to the system $\inputs = \begin{bmatrix} \accel^\top & \angrates^\top \end{bmatrix}^\top$, where $\accel$ is specific acceleration and $\angrates$ is the angular velocity.

\begin{equation} \label{eq:dynamics}
    \boldsymbol{f}(\state, \inputs) = \dot{\state} =
    \begin{bmatrix}
    \Rtheta\vel \\ 
    \Rtheta^\top\begin{bmatrix} 0 & 0 & g \end{bmatrix}^\top + \accel + \vel\times\angrates \\
    S(\att)\angrates \\
    \boldsymbol{0}_{3\times1}
    \end{bmatrix}.
\end{equation}
Since the actual inputs $\accel$ and $\angrates$ are unavailable, we use the IMU to approximate these as $\accel \approx \boldsymbol{y}_\text{accel}$ and $\angrates \approx \boldsymbol{y}_\text{gyro} - \gyrobias$.
These are assumed to be the instantaneous measurement of the specific acceleration and angular rates and are held constant over the time step between estimates.
The rotation matrix $\Rtheta$ transforms vectors in the body-frame into the inertial frame according to the current attitude estimate $\att$.

The Jacobian of the dynamics with respect to the states is

\begin{equation} \label{eq:prop_matrix}
    A(\state, \inputs) = \begin{bmatrix}
        \zeromatrix & \Rtheta & \frac{\partial\Rtheta \vel}{\partial \att} & \zeromatrix \\
        \zeromatrix & -[\angrates]_\times & \frac{\partial \Rtheta^\top \boldsymbol{g}}{\partial \att}  & \zeromatrix \\
        \zeromatrix & \zeromatrix & \frac{\partial S(\att)\angrates}{\partial \att} & -S(\att) \\
        \zeromatrix & \zeromatrix & \zeromatrix & \boldsymbol{I}_{3\times3} \\
    \end{bmatrix},
\end{equation} 
where
\begin{equation}
    S(\att) = \begin{bmatrix}
        1 & \sin\phi\tan\theta & \cos\phi\tan\theta \\
        0 & \cos\phi & -\sin\phi \\
        0 & \sin\phi\sec\theta & \cos\phi\sec\theta
    \end{bmatrix}
\end{equation}
is the relation from body rates to Euler angle rates and properly accounts for the sequential frames associated with the Euler angles.

Note that $A$ is the continuous-time Jacobian and not the discrete-time Jacobian.
To find the discrete-time Jacobian so it can be used in the EKF's discrete updates, we use a second order approximation of the matrix exponential $e^{AT_s}$, where $T_s$ is the period between state propagation updates in the EKF and is given by
\begin{equation}
    A_d(\state,\inputs) = I + A(\state, \inputs)T_s + A(\state, \inputs)^2\frac{T_s^2}{2}.
\end{equation}
 
The propagation step is split into $N$ updates.
This reduces linearization errors that could be introduced.
The state propagation step updates the following $N$ times,
\begin{align}
    \state^-_{k+1,i} &= \state_k + f(\state, \inputs) \frac{T_s}{N}\\
    P_{k+1}^- = A_d(\state, \inputs) P_k &A_d(\state, \inputs)^\top + (Q + GQ_uG^\top)\big(\frac{T_s}{2N}\big)^2,
\end{align}
and the most up-to-date $\state$ gets used in the calculation of $f$ and $A_d$.
The process noise is denoted as $Q$, the uncertainty on the inputs is given as $Q_u$ and the Jacobian of $f$ with respect to the inputs is given as $G$.
$P$ is the covariance of the state estimate.

\subsubsection{Measurement Updates}
The measurements $\boldsymbol{z}$, are incorporated into the estimate of the state $\state$ using the Kalman update based on the predicted measurement $h(\state, \boldsymbol{u})$:
\begin{equation}
    \state_{k+1} = \state_{k} + K(z-h(\state_{k}, \boldsymbol{u}_{k}))
\end{equation}

The general measurement update equation used in the ROScopter EKF is factored into Joseph's form to ensure that the covariance remains positive definite.
It is calculated as,

\begin{align}
    S &= R + CP^-C^\top \\
    K &= P^-C^\top S^{-1} \\
    P^+ &= (I - KC)P^-(I-KC)^\top + LRL^\top, 
\end{align}
where $S$ denotes the uncertainty on the innovation $s = z-h$, $K$ is the Kalman gain and $C$ is the Jacobian of the measurement model with respect to the state, also called the observation Jacobian.

If sensor measurement $z$ is a function of the states and the sensor reading then the covariance of the innovation $S$ will have more terms, in which case,
\begin{equation} \label{eq:full_S}
    S = FRF^\top + GP^-G^\top + CP^-C^\top - 2GP^-C^\top,
\end{equation}
where $F$ is the Jacobian of $z$ with respect to the measurement, $G$ is the Jacobian of $z$ with respect to the estimated state.
As will be seen later, this is useful when the measurement model would otherwise be complex.

Beyond the IMU, the estimator utilizes the following sensors: a barometer for measuring altitude, a magnetometer for measuring heading, and GNSS for measuring position and velocity.

The measurement model $h$, and observation Jacobian $C$, for each sensor are given below.

\paragraph{Barometer} The barometer measures the static atmospheric pressure and uses the model,
\begin{equation}
    h_{\text{baro}} = -\rho g p_d
\end{equation}
where $g$ is the acceleration due to gravity, $p_d$ is the estimated down position and $\rho$ is the air density calculated from the absolute altitude above sea level and the 1976 standard atmosphere model for the troposphere \cite{1976_atmosphere}.
The model yields the expected sensor value for atmospheric pressure at the current altitude above sea level, using GNSS altitude.
Taking the derivative with respect to $\state$, we obtain the observation Jacobian,
\begin{equation}
    C_{\text{baro}} =
    \begin{bmatrix}
        0 & 0 & -\rho g &
        \boldsymbol{0}_{1\times3} &
        \boldsymbol{0}_{1\times3} &
        \boldsymbol{0}_{1\times3} &
    \end{bmatrix}.
\end{equation}

\paragraph{Magnetometer} The magnetometer measures the intensity of the magnetic field in three axes.
This is used to create a measurement of the heading using a tilt-compensated magnetometer model \cite{tilt_mag},
\begin{equation}\label{eq:tilt-mag}
    z_{\text{mag}} = \text{atan2}(\begin{bmatrix} 0&1&0 \end{bmatrix}R_b^{v_1}\boldsymbol{m},\, \begin{bmatrix} 1&0&0 \end{bmatrix}R_b^{v_1}\boldsymbol{m}) .
\end{equation}

Eq. \eqref{eq:tilt-mag} finds the components of the magnetic field $\boldsymbol{m}$ in the horizontal plane by projection using $R_b^{v_1}$ to express the magnetic field measured in the yawed inertial frame $v_1$, and then finds the angle of that projection relative to true north. 

The measurement model is simply
\begin{equation}
    h_{\text{mag}} = \psi,
\end{equation}
where $\psi$ is the estimated yaw of the aircraft.

The observation Jacobian is then,
\begin{equation}
    C_{\text{mag}} =
    \begin{bmatrix}
        \boldsymbol{0}_{1\times3} &
        0 & 0 & 1 &
        \boldsymbol{0}_{1\times3} &
        \boldsymbol{0}_{1\times3} &
    \end{bmatrix}.
\end{equation}
The contribution of the uncertainty of the states used to calculate $z_{\text{mag}}$, is found by calculating $F = \frac{\partial z}{\partial y}$ and $G = \frac{\partial z}{\partial \state}$ and using Eq. \eqref{eq:full_S}.

\paragraph{GNSS} The GNSS update includes measurements of the velocity of the aircraft in the inertial frame and the absolute position in latitude and longitude.
These are collated into one update where

\begin{equation}
    z_{\text{gnss}} = \begin{bmatrix}
        p_{n,gnss} \\
        p_{e,gnss} \\
        v_{n,gnss} \\
        v_{e,gnss} \\
        v_{d,gnss} \\
    \end{bmatrix}.
\end{equation}

The GNSS altitude measurement is omitted because commercially available single-band GNSS antennas can have large drifts in altitude measurement.
The raw latitude and longitude angles of the GNSS are converted to a local north and east position relative to an initial latitude and longitude recorded on initialization, or provided through configuration.
This is done by making a spherical earth assumption:
\begin{align}
    p_{\mathrm{n,gnss}} &= (d_{\text{lat}}-d_{\text{init, lat}})r_{\text{earth}}\\
    p_{\mathrm{e,gnss}} &= (d_{\text{lon}}-d_{\text{init, lon}})r_{\text{earth}}\cos(d_{\text{lat}}),\\
\end{align}
where $r_{\text{earth}}$ is the radius of the Earth.

The measurement model used for GNSS is
\begin{equation}
    h_{\text{gnss}} = \begin{bmatrix}
        p_n \\
        p_e \\
        \Rtheta \vel
    \end{bmatrix},
\end{equation}
taking the estimated north and east positions directly from the state and the body velocities expressed in the inertial frame according to the current attitude estimate.
This allows for tight coupling between the attitude angles and the high certainty velocity measurements from GNSS.
An additional advantage is that the gyroscope biases and the GNSS velocities are tightly coupled.

The measurement model yields the observation Jacobian
\begin{equation}
    C_{\text{gnss}} = \begin{bmatrix}
        \boldsymbol{I}_{2\times2} & \boldsymbol{0}_{2\times4} & \boldsymbol{0}_{2\times3} & \boldsymbol{0}_{2\times3}  \\
        \boldsymbol{0}_{3\times3} & \Rtheta & \frac{\partial \Rtheta\vel}{\partial\att} & \boldsymbol{0}_{3\times3} \\
    \end{bmatrix}
\end{equation}

\subsection{Navigation Stack}
ROScopter's navigation stack consists of a path planner, a path manager, and a path follower, as shown in Figure \ref{fig:roscopter-system-architecture}.
The path planner is responsible for collecting waypoints and publishing them to the path manager.
In the default implementation of ROScopter, the $i$th waypoint $\boldsymbol{w}_i=(\boldsymbol{p}_i, \psi_i)$ is defined by the user and is specified by a desired 3-D position $\boldsymbol{p}_i$ and heading $\psi_i$.

The path manager is responsible for generating a trajectory between consecutive waypoints.
We define each trajectory between waypoints as a \emph{waypoint leg}.
Each trajectory command is a vector of 3-D positions, heading, and their derivatives of the form
\begin{equation}\label{eq:path-manager-desired-trajectory}
    \boldsymbol{u}_\text{traj}(\tau) = [\boldsymbol{p}(\tau), \dot{\boldsymbol{p}}(\tau), \ddot{\boldsymbol{p}}(\tau), \psi(\tau), \dot{\psi}(\tau), \ddot{\psi}(\tau)]^T,
\end{equation}
where $\boldsymbol{u}_\text{traj}$ is the desired trajectory.
The path manager publishes each desired trajectory setpoint, $\boldsymbol{u}_\text{traj}(\tau)$, at a configurable rate, where $\tau \in [0,1]$ is normalized time, given by
\begin{equation}\label{eq:path-manager-tau}
    \tau = \frac{t}{T_\text{path}},
\end{equation}
and where $t$ is the time along the current waypoint leg, and $T_\text{path}$ is the total time it will take to travel the waypoint leg.

In the default implementation of ROScopter, the path manager generates straight-line trajectories between consecutive waypoints $\boldsymbol{p}_i$ and $\boldsymbol{p}_{i+1}$.
Thus, $\boldsymbol{p}(\tau)$ and $\psi(\tau)$ are linear interpolations between the desired positions and headings at consecutive waypoints, where $\boldsymbol{p}(0)=\boldsymbol{p}_i$, $\psi(0)=\psi_i$, $\boldsymbol{p}(1)=\boldsymbol{p}_{i+1}$, and $\psi(1)=\psi_{i+1}$.
The rate of interpolation is given by the path parameter $\sigma(\tau)$, which can be any monotonically increasing function with $\sigma(0)=0$ and $\sigma(1)=1$.
In ROScopter, we define $\sigma(\tau)$ to be a 5th-order smoothstep function, resulting in zero velocity and acceleration at each waypoint, giving a smooth transition between waypoint legs.

The trajectory follower is a simple PID controller on position that incorporates position, velocity, acceleration feedforward terms from the differentially flat formulation of \cite{diff_flat} to follow the trajectory setpoints from Eq. \eqref{eq:path-manager-desired-trajectory}.
As in \cite{diff_flat}, the trajectory follower module produces control setpoints of the form
\begin{equation}\label{eq:traj-follower-output}
    \boldsymbol{u}_\text{angle}=[\phi^d, \theta^d, r^d, T^d]^T,
\end{equation}
where $\boldsymbol{u}_\text{angle}$ is the desired setpoint published to an angle controller and is composed of the desired roll $\phi^d$, pitch $\theta^d$, yaw rate $r^d$, and thrust $T^d$, respectively.
These reference setpoints are then published to the controller.

\subsection{Controller}
\begin{figure}
    \centering
    \includegraphics[width=0.95\linewidth]{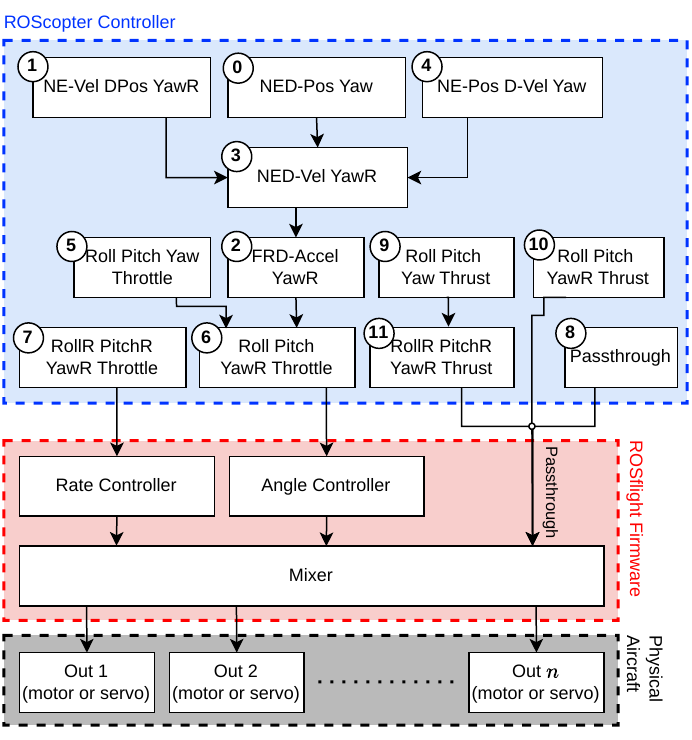}
    \caption{Entry points for the ROScopter controller. Each entry point is a PID controller and is described in Table \ref{tab:roscopter-cascading-controller}. The ROScopter controller sends commands to the ROSflight firmware.}
    \label{fig:roscopter-cascading-controller}
\end{figure}

The ROScopter controller is responsible for taking in upstream controller commands and converting them to low-level commands.
Because different applications that feed into the controller (e.g. the ROScopter trajectory follower) produce different reference commands, the ROScopter controller allows users to select a variety of input reference setpoint types as shown in Figure \ref{fig:roscopter-cascading-controller}.
Each box shown in Figure \ref{fig:roscopter-cascading-controller} is a separate PID controller and is described in Table \ref{tab:roscopter-cascading-controller}.
In this formulation, users can send reference commands to any of the entry points in the ROScopter controller, and the commands cascade down to the lowest-level control loops.
For example, under a particular navigation scheme, a user may send reference inertial velocities to the ROScopter controller.
This velocity loop is closed by controller 3, which returns accelerations in the vehicle-1 frame (see \cite{uavbook}).
These reference accelerations are passed to controller 2, and then to controller 6, where the resulting reference angle commands are published to the ROSflight firmware.

\begin{table}
    \centering
    \begin{tabular}{@{}p{0.03\columnwidth}|p{0.38\columnwidth}>{\hangindent=0.1in}p{0.5\columnwidth}@{}}
         & Name & Description of reference commands \\
         \toprule
         0  & NED-Pos Yaw & Inertial NED positions and yaw \\
         1  & NE-Vel D-Pos YawR & Inertial N-E velocities, D position, and yaw rate \\
         2  & FRD-Accel YawR & Vehicle-1 (front-right-down) accelerations and yaw rate \\
         3  & NED-Vel YawR & Inertial NED velocities and yaw rate \\
         4  & NE-Pos D-Vel Yaw & Inertial N-E velocities and yaw \\
         5  & Roll Pitch Yaw Throttle & Roll, pitch, yaw, and throttle \\
         6  & Roll Pitch YawR Throttle & Roll, pitch, yaw rate, and throttle \\
         7  & RollR PitchR YawR Throttle & Roll rate, pitch rate, yaw rate throttle \\
         8  & Pass-through & Pass-through to ROSflight mixer \\
         9 & Roll Pitch Yaw Thrust & Roll, pitch, yaw, and thrust to pass-through \\
         10 & Roll Pitch YawR Thrust & Roll, pitch, yaw rate, and thrust to pass-through \\
         11 & RollR PitchR YawR Thrust & Roll rate, pitch rate, yaw rate, and thrust to pass-through \\
         \bottomrule
    \end{tabular}
    \caption{Description of each entry point into the ROScopter cascading controller architecture.}
    \label{tab:roscopter-cascading-controller}
\end{table}
Many different types of control reference inputs are standard in multirotor applications, so this formulation of ROScopter prevents users from needing to write a low-level controller for all different types of standard reference commands.
Additionally, a given input type is selected by setting the appropriate mode as each command message is published, so different control input types may be used arbitrarily.
For example, the default configuration of ROScopter uses controller 4 during takeoff and controller 10 during other portions of flight.
Controller 10 takes in reference setpoints computed by the trajectory follower, as shown in Eq. \eqref{eq:traj-follower-output}.

ROScopter is designed to work with a flight controller unit (FCU) running ROSflight firmware and thus returns commands in one of the three modes that ROSflight accepts, as shown in Figure \ref{fig:roscopter-cascading-controller}.
As described in \cite{rosflight2025}, the ROSflight firmware contains two low-level controllers, an angle controller and a rate controller.
Additionally, ROSflight allows commands to bypass all low-level loops on the firmware and progress directly to the mixer, called pass-through mode.
Thus, the ROScopter controller publishes one of the following command vectors
\begin{align}
    \boldsymbol{u}_\text{angle} &= [0, 0, \delta_t^d, \phi^d, \theta^d, r^d, \boldsymbol{0}_{1\times4}]^T, \label{eq:controller-angle-output} \\
    \boldsymbol{u}_\text{rate} &= [0, 0, \delta_t^d, p^d, q^d, r^d, \boldsymbol{0}_{1\times4}]^T, \label{eq:controller-rate-output} \\
    \boldsymbol{u}_\text{pass-through} &= [0, 0, T_z^d, Q_x^d, Q_y^d, Q_z^d, \boldsymbol{0}_{1\times4}]^T, \label{eq:controller-pass-through-output}
\end{align}
where $\phi^d$, $\theta^d$, are the desired roll and pitch, $p^d$, $q^d$, $r^d$ are the desired roll rate, pitch rate, and yaw rate, $\delta_t^d \in [0,1]$ is the desired throttle setpoint, and $Q_x^d$, $Q_y^d$, $Q_z^d$, $T_z^d$ are the desired roll torque, pitch torque, yaw torque, and thrust in the body $z$ direction, respectively. 
The zeros in these command messages are to comply with the default interpretation of this $\boldsymbol{u}$ command vector, described in \cite{rosflight2025}.

\section{Tutorial}\label{tutorial}
This section offers a brief tutorial on how users can expect to use the ROScopter autonomy stack to accelerate testing and development.
This is not intended to be a comprehensive tutorial, but is intended to describe examples of how ROScopter can be customized and adapted to fit researcher needs.
See the project website for the most up-to-date information and tutorials using ROScopter.

\subsection{Default ROScopter behavior}
This section describes ROScopter's out-of-the-box functionality. 
By default, the ROScopter autonomy stack allows multirotor vehicles to follow waypoints defined by a 3-D world coordinate and a heading.
The 3-D position can be specified using latitude, longitude and altitude (LLA) coordinates or meters north, east, and down from an origin.
Paths between these waypoints are planned as straight line segments and the heading is linearly interpolated between waypoints.

Although this simple functionality will not satisfy all researchers' needs, the default ROScopter autonomy stack is useful in many applications.
Applications such as search-and-rescue may only require that a multirotor travel to a particular sequence of waypoints; in this case, ROScopter could effectively sit underneath application code.

\subsection{Customizing ROScopter}
The default behavior described in the previous section will not satisfy all user's needs.
Users that need to access the low-level loops and estimation schemes of ROScopter will likely need to customize ROScopter.
This section describes how ROScopter was designed for customizability and modularity.

Every module in Figure \ref{fig:roscopter-system-architecture} is implemented as a ROS 2 node with clearly defined ROS 2 interfaces (e.g. publishers, subscribers, and service calls).
Additionally, as discussed in Section \ref{system-architecture:code}, each module is implemented using an inheritance pattern to facilitate modification.
This reduces dependencies between modules, isolates functionality, and allows for easy customization of a particular module without modifying any other modules, thus lowering the barrier to entry.
For example, a different state estimator with different measurement or update functions could be implemented by rewriting the existing estimator module or by sub-classing the estimator interface.
As long as the ROS~2 interfaces between the estimator and all other modules remain the same (in this case, the state estimate message), then the other modules are agnostic to how the state estimate values are computed, and the new estimator code will integrate seamlessly into ROScopter.

The dependency on ROS 2 also enables easy runtime modification of the autonomy stack.
Each module is a separate executable, so different configurations of the autonomy stack can be selected at runtime.
For example, if a new estimator were implemented into ROScopter, users could easily swap back to the default version of the estimator without recompiling by simply launching the default estimator node.
This behavior can be useful when comparing several methods side by side, as is common in research.
Runtime behavior can also be easily configured using ROS 2 parameters, which allow users to dynamically adjust parameters (e.g. controller gains, estimator sensitivities, path parameters, etc.) without needing to restart or recompile a given node.
This results in a powerful testing framework that allows for quicker and easier simulation and hardware testing of new research.

Often, users may need to replace groups of the default nodes.
For example, if a user wants to plan spline paths instead of straight lines, then both the path planner and path manager may need to be removed or modified.
The linear, cascaded architecture of ROScopter makes this easy; as long as the ROS 2 interfaces between the new modules and modules lower in the cascaded architecture remain the same, then the new modules will interface seamlessly into ROScopter's architecture.

\subsection{Intended workflow for ROScopter}
When using the ROScopter autonomy stack, users should first determine which nodes should be replaced or modified given a particular application.
Application code can then be developed and integrated into the autonomy stack as described in the previous sections.
Simulation tests can easily be run by configuring the autonomy stack at runtime and running application code instead of the default ROScopter modules.

ROScopter relies on the ROSflight simulation environment \cite{rosflight2025}.
The default multirotor simulation environment makes several assumptions about the aircraft including mass, vehicle size, and location and number of motors.
If no hardware tests will be performed, then these values can be left as their default values.
If hardware tests will be performed, then setting these values appropriately will result in a more realistic simulation environment and an easier sim-to-real transition.
Since the ROSflight simulation environment uses the exact same code between simulation and hardware environments, no major changes to the application code will need to be changed when transitioning from simulation to hardware tests.
However, as every simulation environment is different than the real world, small changes to controller gains or other tuning parameters may need to be adjusted when first flying in hardware.

\section{Results}\label{results}
The performance of the default ROScopter autonomy stack was validated both in simulation and hardware environments.

\subsection{Waypoint-path following}
To demonstrate the basic functionality of ROScopter, a simple waypoint mission was flown in simulation and hardware.
This waypoint mission consisted of three waypoints, shown in Table \ref{tab:waypoint-mission}.
These hardware results used a HolyBro x650 quadcopter frame with the autopilot configuration 2 as described in \cite{rosflight2025}, which uses the Varmint flight control unit with an integrated Jetson Orin developed by AeroVironment, Inc.
The ROScopter path manager was configured to have a maximum path velocity of 3 m/s between waypoints.
Command setpoints were sent from ROScopter over a serial connection to the ROSflight mixer in pass-through mode \cite{rosflight2025}, meaning that all control loops were closed on the companion computer.

\begin{table}
    \centering
    \begin{tabular}{c|ccc}
         Waypoint & $w_1$ & $w_2$ & $w_3$ \\
         \toprule
         North (m) & 0 & -20 & -20 \\
         East (m)  & 0 & 0 & 20 \\
         Down (m)  & -5 & -8 & -5 \\
         Heading (deg) & 130 & 130 & 130 \\
         \bottomrule
    \end{tabular}
    \caption{Desired NED positions and headings of the waypoint mission shown in Figure \ref{fig:experiments-roscopter-default-functionality}.}
    \label{tab:waypoint-mission}
    \vspace{-15pt}
\end{table}

\begin{figure}
    \centering
    \includegraphics[width=0.9\columnwidth]{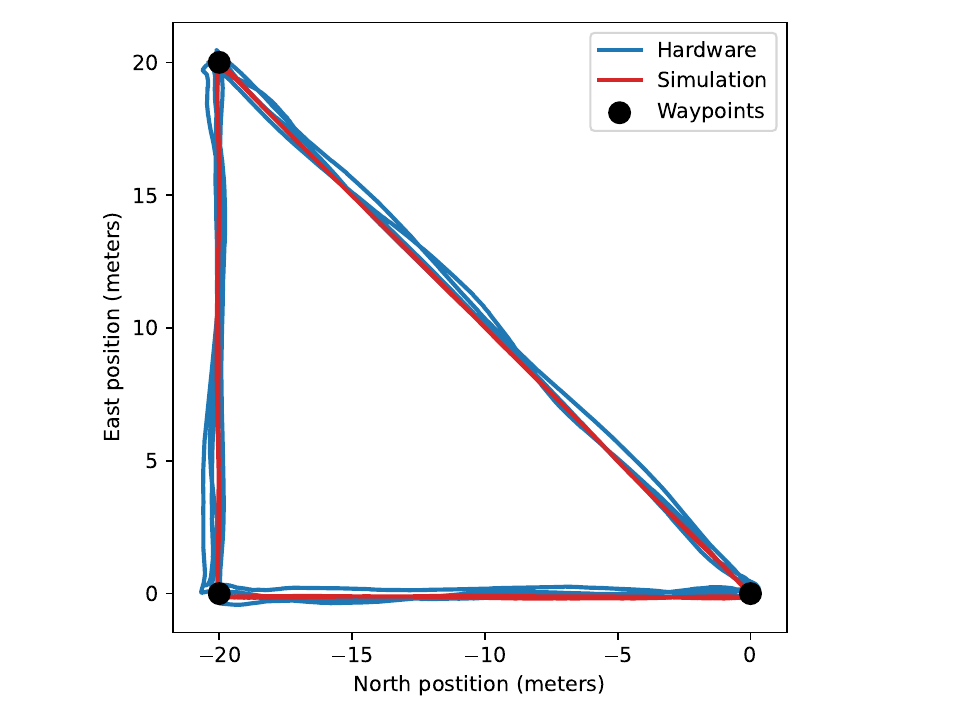}
    \caption{Top-down view of simulation and hardware flight test results using the exact same control and estimation gains on a HolyBro x650 quadcopter frame. Simulation and hardware responses match, highlighting how ROScopter can accelerate simulation to hardware transitions.}
    \label{fig:experiments-roscopter-default-functionality}
    \vspace{-15pt}
\end{figure}

\begin{figure}
    \centering
    \includegraphics[width=0.9\columnwidth]{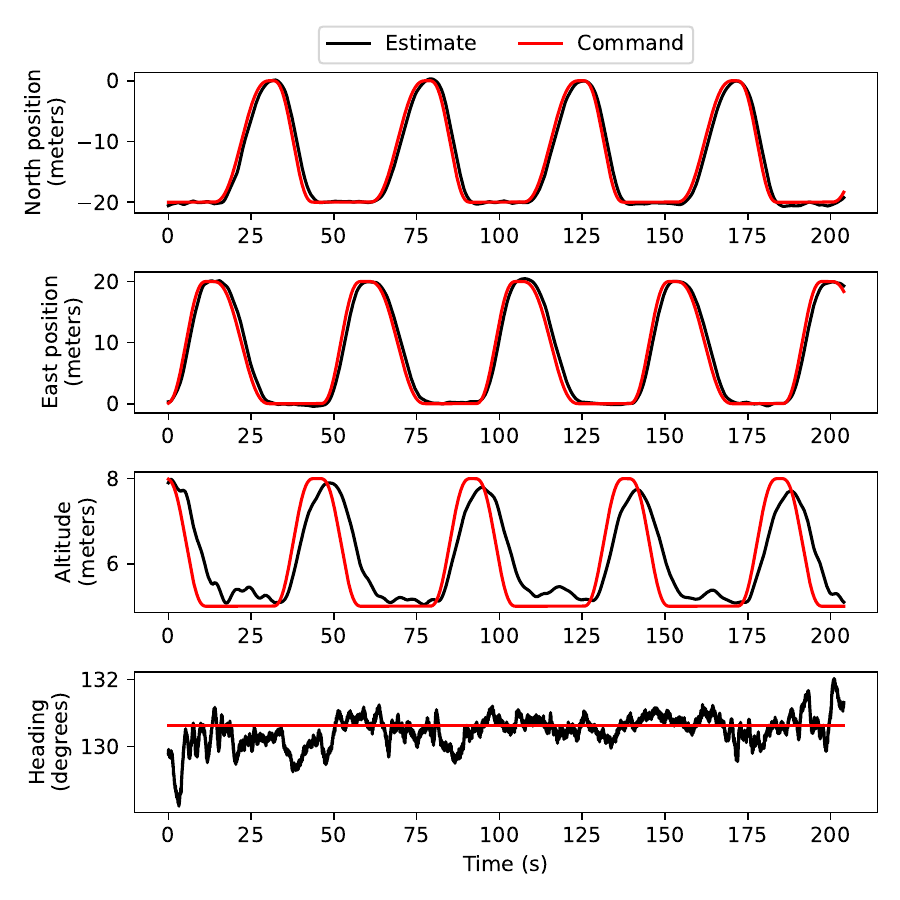}
    \caption{Response of the vehicle using ROScopter to commanded position and heading setpoints in hardware for the same trajectory shown in Figure \ref{fig:experiments-roscopter-default-functionality}.}
    \label{fig:experiments-roscopter-default-functionality-split}
    \vspace{-15pt}
\end{figure}

The results from the simulation and hardware experiments are shown in Figures \ref{fig:experiments-roscopter-default-functionality} and \ref{fig:experiments-roscopter-default-functionality-split}.
Mild wind was present during the hardware flight test while no wind was simulated.
Both experiments (in simulation and hardware) used the exact same autonomy stack with identical parameter configuration and control gains.
To arrive at these control gains, the system was first tuned in simulation, and only minor adjustments were made when transitioning to hardware.

\begin{figure}
    \centering
    \includegraphics[width=\columnwidth]{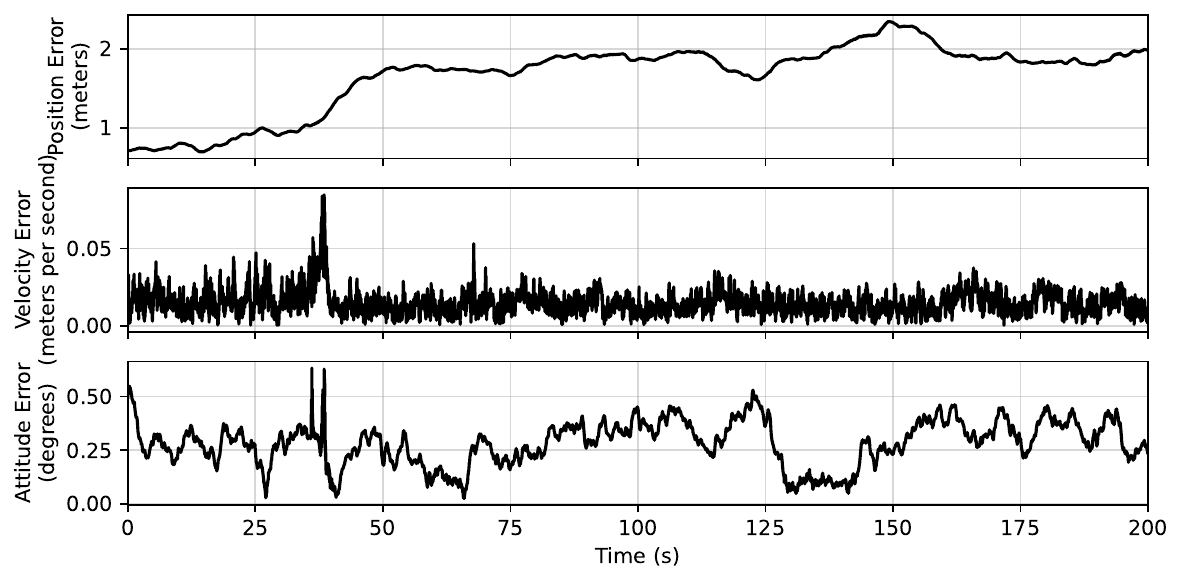}
    \caption{Errors between truth and estimated states for ROScopter's estimator from simulation.}
    \label{fig:est-vs-sim}
\end{figure}

These results show that ROScopter is able to perform basic waypoint following missions out-of-the-box.
Furthermore, the simulation and hardware results align when using the same control gains and parameter configurations.
This shows that ROScopter is able to accelerate the transition from simulation to hardware since performance between simulation and hardware is comparable.
Additionally, this shows that the vast majority of controller gain tuning can be performed in simulation, requiring only slight modifications when transitioning to hardware.
The error in estimated state in simulation is shown in Figure \ref{fig:est-vs-sim}, with RMS error of 1.72 meters in position, 0.017 meters per second in velocity and 0.301 degrees in attitude.

\subsection{Comparison to PX4}
We now compare the performance of ROScopter to a state-of-the-art open-source autopilot, PX4 \cite{px4}.
Since ROScopter is not designed to be directly comparable to PX4 (see Section \ref{related-work}), the purpose of these tests is to validate that ROScopter's basic functionality meets the needs of researchers.

\begin{figure}
    \centering
    \includegraphics[width=0.9\columnwidth]{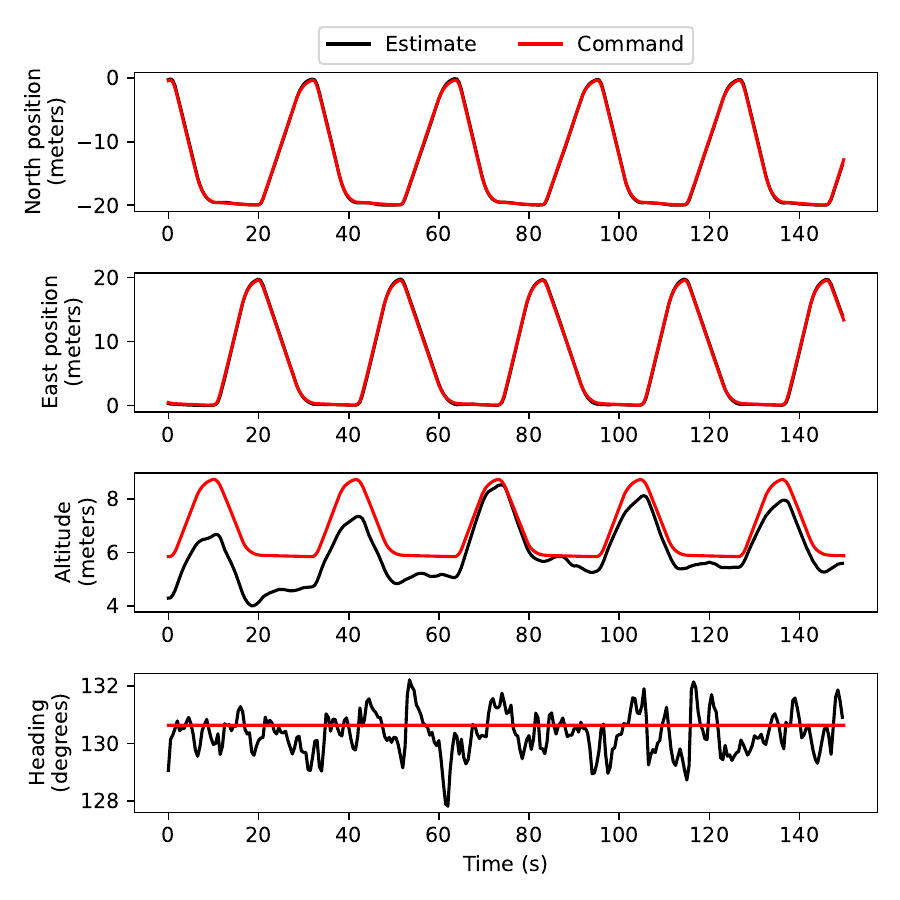}
    \caption{Response of the vehicle using PX4 in hardware when flying the same waypoint path shown in Figure \ref{fig:experiments-roscopter-default-functionality}.}
    \label{fig:experiments-px4-default-functionality-split}
    \vspace{-15pt}
\end{figure}

To compare the closed-loop controller response for waypoint-following missions, a Pixracer Pro FCU was flashed with ROSflight firmware and mounted to a HolyBro x650 frame.
For these experiments, a Raspberry Pi 5 was used as the companion computer (configuration 1 as described in \cite{rosflight2025}.
A mission was loaded to the ROScopter path planner and was flown.
The Pixracer Pro was flashed with PX4 firmware version v1.15.4 and mounted to the same HolyBro x650 frame.
The same mission was loaded using a laptop with the QGroundControl software, and was flown.
Figure \ref{fig:experiments-roscopter-default-functionality-split} shows ROScopter's closed-loop performance when following the same trajectory shown in Figure \ref{fig:experiments-roscopter-default-functionality}.
Figure \ref{fig:experiments-px4-default-functionality-split} shows PX4's closed-loop performance when following this same trajectory.
Table \ref{tab:px4-vs-rf-rmse} shows the root-mean-squared errors between each autopilot's estimated state and the ideal straight-line path between waypoints.
We expect that both responses could be improved with more rigorous gain tuning.
While PX4 contains many additional features, excellent autopilot functionality, and improved performance, these results show that ROScopter offers similar performance for basic waypoint-following functionality while maintaining a lean, understandable, modular, and clean code-base.

\begin{table}
    \centering
    \begin{tabular}{c|ccc|c}
         & X (m) & Y (m) & Z (m) & Total (m) \\
         \toprule
         ROScopter & 0.197 & 0.145 & 0.474 & 0.533 \\
         PX4       & 0.152 & 0.102 & 0.431 & 0.469 \\
         \bottomrule
    \end{tabular}
    \caption{RMS errors between estimated state and desired straight-line waypoint path from hardware flight tests. PX4's RMSE only includes data after $t=63$s, after altitude control had stabilized.}
    \label{tab:px4-vs-rf-rmse}
    \vspace{-15pt}
\end{table}

\begin{figure}
    \centering
    \includegraphics[width=0.9\columnwidth]{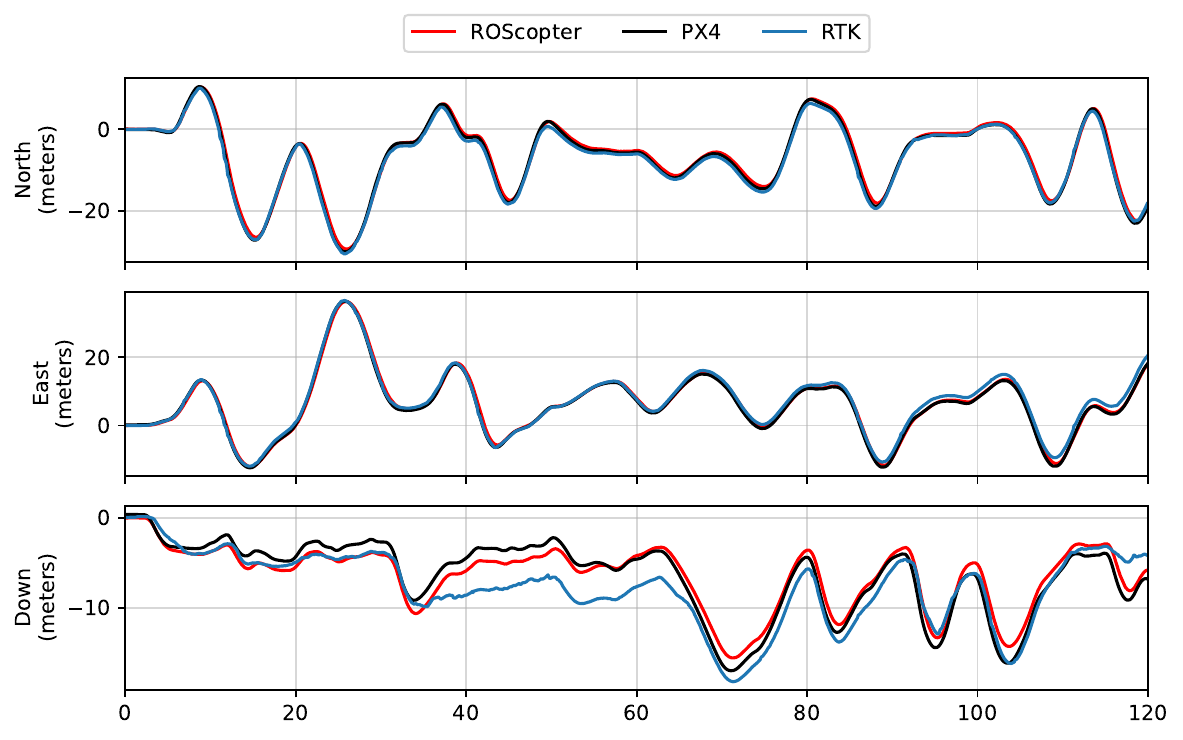}
    \includegraphics[width=0.9\columnwidth]{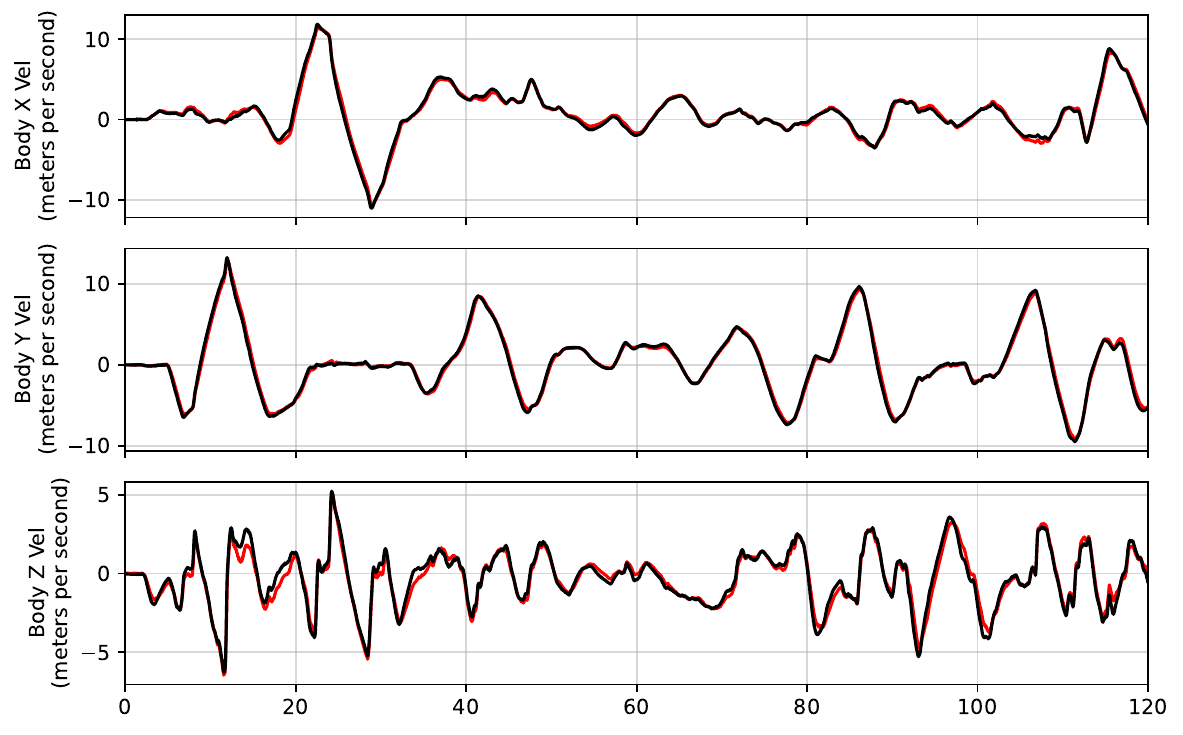}
    \includegraphics[width=0.9\columnwidth]{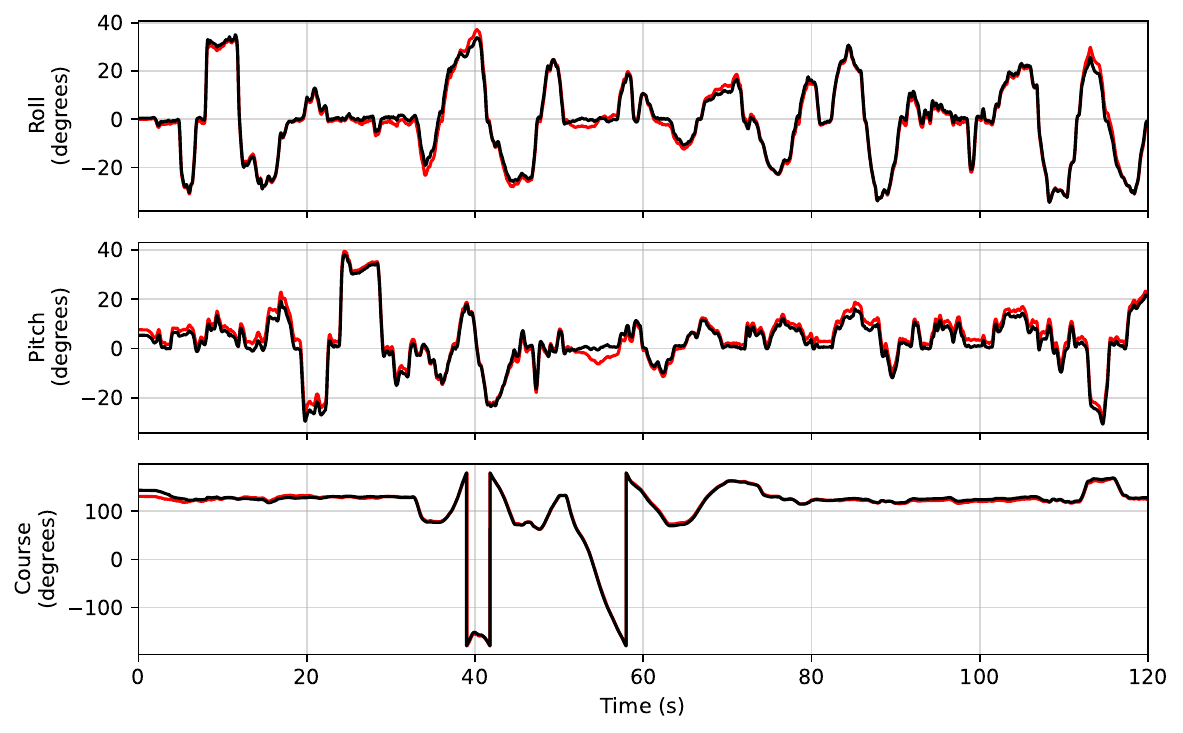}
    \caption{Comparison of the PX4 and ROScopter estimator output given the same sensor information. This data was collected under PX4, with RTK positional data recorded.}
    \label{fig:experiments-px4-vs-rosflight-estimator}
    \vspace{-15pt}
\end{figure}

To show that ROScopter's estimation is satisfactory, an in-flight comparison of estimates to PX4 was conducted.
The configuration from above was flown, performing various maneuvers under manual control.
The sensor data and PX4 estimates were recorded at a high rate.
ROScopter's estimator was then run on the resulting sensor data and compared to PX4's estimate.
Figure \ref{fig:experiments-px4-vs-rosflight-estimator} shows the comparison, and demonstrates that the ROScopter estimator performs similarly to PX4's industry-standard estimator.
Positional data was recorded via RTK GNSS to serve as ground truth, but was not used by either the PX4 or ROScopter estimators.
Attitude ground truth was unavailable in the outdoor setting.

\section{Conclusion}\label{conclusion}
In this paper we presented ROScopter, a lean researcher-focused autopilot for multirotor vehicles.
ROScopter is designed to lower the barrier to entry to UAV research by providing a simple, customizable autopilot with clear, extensible code and complete documentation.
ROScopter also accelerates testing application code by enabling the same autonomy stack that runs in simulation to control the vehicle in hardware, with no changes.
ROScopter is based heavily on ROS 2 to enhance modularity and customizability.
Simulation and hardware tests show that ROScopter is able to perform basic autopilot functionality at a similar level to state-of-the-art autopilots while maintaining a significantly reduced code base.

\bibliographystyle{IEEEtran}
\bibliography{bibi}

\end{document}